\def\year{2021}\relax
\documentclass[letterpaper]{article} 
\usepackage{aaai21}  
\usepackage{times}  
\usepackage{helvet} 
\usepackage{courier}  
\usepackage[hyphens]{url}  
\usepackage{graphicx} 
\usepackage{natbib}  
\usepackage{caption} 
\usepackage{times}
\usepackage{latexsym}
\usepackage{amsmath}
\usepackage{url}  
\usepackage{multicol}
\usepackage{color}
\usepackage{multirow}
\usepackage{algorithm}            
\usepackage{algpseudocode}
\usepackage{amsfonts}
\usepackage{tabularx}
\usepackage{stackengine}
\usepackage{subcaption}
\usepackage[switch]{lineno}

\makeatletter

\newcommand{\multiline}[1]{%
  \begin{tabularx}{\dimexpr\linewidth-\ALG@thistlm}[t]{@{}X@{}}
    #1
  \end{tabularx}
}

\newcommand{\printfnsymbol}[1]{%
  \textsuperscript{\@fnsymbol{#1}}%
}

\makeatother

\title{RpBERT: A Text-image Relation Propagation-based BERT Model for Multimodal NER}

\author {
Lin Sun\textsuperscript{\rm 1}\thanks{Equal contribution.},
 Jiquan Wang\textsuperscript{\rm 2,\rm 1}\printfnsymbol{1},
 Kai Zhang\textsuperscript{\rm 3},
Yindu Su\textsuperscript{\rm 2,\rm 1},
 Fangsheng Weng\textsuperscript{\rm 1}
 \\
}
\affiliations {
    \textsuperscript{\rm 1} Department of Computer Science, Zhejiang University City College, Hangzhou, China\\
    \textsuperscript{\rm 2} College of Computer Science and Technology, Zhejiang University, Hangzhou, China\\
		 \textsuperscript{\rm 3} Department of Computer Science and Technology, Tsinghua University, Beijing, China\\
    sunl@zucc.edu.cn, wangjiquan@zju.edu.cn, drogozhang@gmail.com, yindusu@zju.edu.cn
}

\begin{document}
 
	
\maketitle

\begin{abstract}
Recently multimodal named entity recognition (MNER) has utilized images to improve the accuracy of NER in tweets.
However, most of the multimodal methods use attention mechanisms to extract visual clues regardless of whether the text and image are relevant.
Practically, the irrelevant text-image pairs account for a large proportion in tweets.
The visual clues that are unrelated to the texts will exert uncertain or even negative effects on multimodal model learning.
In this paper, we introduce a method of text-image relation propagation into the multimodal BERT model.
We integrate soft or hard gates to select visual clues and propose a multitask algorithm to train on the MNER datasets.
In the experiments, we deeply analyze the changes in visual attention before and after the use of text-image relation propagation.
Our model achieves state-of-the-art performance on the MNER datasets.
The source code is available online\footnote{\url{https://github.com/Multimodal-NER/RpBERT}}.

\end{abstract}

\section{Introduction}

\begin{figure}[tb]
\begin{minipage}[t]{0.5\textwidth}
\centering
\includegraphics[width=0.6\textwidth]{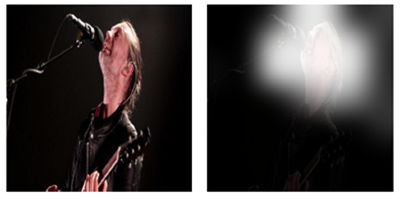}
\small 
\begin{tabular}{c}
(a) \textbf{[PER Radiohead]} offers old and new at first concert  in \\  four years.
\end{tabular}
\end{minipage}
\\
\begin{minipage}[t]{0.5\textwidth}
\centering
\includegraphics[width=0.6\textwidth]{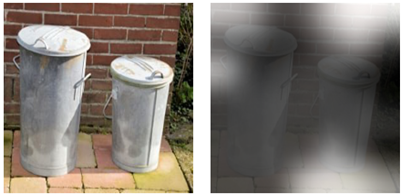}
\small
\begin{tabular}{c}
(b) Nice image of \textbf{[PER Kevin Love]} and \textbf{[PER Kyle Korver]} \\
during 1st half \#NBAFinals \#Cavsin9 \# {\color{red}\textbf{[LOC Cleveland]}.}
\end{tabular}
\end{minipage}
\caption{Visual attention examples of MNER from \cite{lu2018visual}. The left column is a tweet's image and the right column is its corresponding attention visualization.  (a) Successful case, (b) failure case.}\label{fig:visAttExample}
\end{figure}

Social media platforms such as Twitter have become part of the everyday lives of many people.
They are important sources for various information extraction applications such as open event extraction~\cite{wang2019open} and social knowledge graph construction~\cite{hosseini2019implicit}.
As a key component of these  applications, named entity recognition (NER) aims to detect named entities (NEs) and classify them into predefined types, such as person (PER), location (LOC) and organization (ORG).
Recent works on tweets based on multimodal learning have been increasing~\cite{moon2018multimodal,lu2018visual,zhang2018adaptive,arshad2019aiding,yu2020improving}.
These researchers investigated to enhance linguistic representations with the aid of visual clues in tweets.
Most of the MNER methods used attention weights to extract visual clues related to the NEs~\cite{lu2018visual,zhang2018adaptive,arshad2019aiding}.
For example, Fig.~\ref{fig:visAttExample}(a) shows a successful visual attention example in~\cite{lu2018visual}.
In fact, texts and images in tweets could also be irrelevant.
Vempala and Preo{\c{t}}iuc-Pietro~\shortcite{vempala2019categorizing} categorized  text-image relations according to  whether the ``Image adds to the tweet meaning''.
The ``Image does not add to the tweet meaning''  type accounts for approximately 56\% of instances in Vempala's text-image relation classification (TRC) dataset.
In addition, we trained a classifier of whether the ``Image adds to the tweet meaning'' on a large randomly collected corpus, Twitter100k~\cite{hu2017twitter100k},  and the proportion of  classified negatives was approximately 60\%.
The attention-based models would also produce visual attention although the text and image are irrelevant, 
and such visual attention might exert negative effects on the text inference.
Fig.~\ref{fig:visAttExample}(b) shows a failure visual attention example.
The visual attention focuses on the wall and ground, resulting in tagging ``[ORG] Cleveland'' with the wrong label ``LOC''.

%

In this paper, we consider inferring the text-image relation  to address the problem of inappropriate visual attention clues in multimodal models. 
The contributions of this paper can be summarized as follows:
\begin{itemize}
\item 
We propose a novel text-image relation propagation-based multimodal BERT model.
We investigate the soft and hard ways of propagating text-image relations through the model by training.  
A training procedure for the multiple tasks of text-image relation classification and  downstream NER  is also presented.
%
\item 
We provide insights into the visual attention by numerical distributions and heat maps.
Text-image relation propagation can not only reduce the interference from irrelevant images but also leverage more visual information for relevant text-image pairs.
\item 
The experimental results show that the failure cases in the related works are correctly recognized by our model, and the state-of-the-art performance is achieved in this paper.

%
%
\end{itemize}



\section{Related Work}

\smallskip
\noindent
\textbf{Multimodal NER} 
Moon et al.~\shortcite{moon2018multimodal} proposed a modality-attention module at the input of an NER network.
The module computed a weighted modal combination of word embeddings, character embeddings, and visual features.
Lu et al.~\shortcite{lu2018visual} presented a visual attention model to find the image regions related to the content of the text. 
The attention weights of the image regions were computed by a linear projection of the sum of the text query vector and  regional visual representations. 
The extracted visual context features were incorporated into the word-level outputs of the biLSTM model.
Zhang et al.~\shortcite{zhang2018adaptive} designed an adaptive co-attention network (ACN) layer, which was between the LSTM and CRF layers.
The ACN contained a gated multimodal fusion module to learn a fusion vector of the visual and linguistic features.
The author designed a filtration gate to determine whether the fusion feature was helpful in improving the tagging accuracy of each token.
The output score of the filtration gate was computed by a sigmoid activation function.
Arshad et al.~\shortcite{arshad2019aiding} also presented  a gated multimodal fusion representation for each token.
The gated fusion was a weighted sum of the visual attention feature and token alignment feature.
The visual attention feature was calculated by the weighted sum of VGG-19~\cite{simonyan2014very} visual features and the weights were the additive attention scores between a word query and image features.
Overall, the problem of the attention-guided models is that the extracted visual contextual cues do not match the text for irrelevant text-image pairs.
The authors of \cite{lu2018visual,arshad2019aiding} showed failed examples in which unrelated images provided  misleading visual attention and yielded prediction errors.


\smallskip
\noindent
\textbf{Pretrained multimodal BERT} 
The pretrained model BERT has achieved great success in natural language processing (NLP). 
The latest presented visual-linguistic models based on the BERT architecture include VL-BERT~\cite{su2019vl}, ViLBERT~\cite{lu2019vilbert}, VisualBERT~\cite{li2019visualbert}, UNITER~\cite{chen2020uniter}, LXMERT~\cite{tan2019lxmert},  and Unicoder-VL~\cite{li2020unicoder}.
We summarize and compare the existing visual-linguistic BERT models in three aspects as follows: 
1) \textbf{Architecture}. The structures of Unicoder-VL, VisualBERT, VL-BERT, and UNITER were the same as that of vanilla BERT. The image and text tokens  were combined into a sequence and fed into BERT to learn contextual embeddings.
LXMERT and ViLBERT separated visual and language processing into two streams that interacted through cross-modality or co-attentional transformer layers respectively.
2) \textbf{Visual representations}. The image features could be represented as region-of-interest (RoI) or block regions.
All the above pretrained models used Fast R-CNN~\cite{girshick2015fast} to detect objects and pool RoI features.
The purpose of RoI detection is to reduce the complexity of visual information and perform the task of masked region classification with  linguistic clues~\cite{su2019vl,li2020unicoder}.
However, for the irrelevant text-image pairs, the non-useful and salient visual features could increase the interference with the linguistic features.
Moreover, object recognition categories are limited and many NEs have no corresponding object class, such as company trademark and scenic location.
3) \textbf{Pretraining tasks}. The models were trained on image caption datasets such as the COCO caption dataset~\cite{chen2015microsoft} or Conceptual Captions~\cite{sharma-etal-2018-conceptual}.
The pretraining tasks mainly include  masked language modeling (MLM), masked region classification (MRC)~\cite{chen2020uniter,tan2019lxmert,li2020unicoder,su2019vl}, and image-text matching (ITM)~\cite{chen2020uniter,li2020unicoder,lu2019vilbert}.
The ITM task is a binary classification, which defines the pairs in the caption dataset as positives and the pairs generated by replacing the image or text in a paired example with other randomly selected samples as negatives.
It assumed that the text-image pairs in the caption datasets were highly related; however, this assumption could not be established in the text-image pairs of tweets.


\smallskip
Visual features are always directly concatenated with linguistic features~\cite{yu2019adapting} or extracted by attention weights in the latest multimodal models, regardless of whether the images contribute to the semantics of the texts, resulting in failed MNER examples shown in Table~\ref{tab:case}.
Therefore, in this work, we explore a multimodal variant of BERT to perform MNER for tweets with different text-image relations.






%



%
%
%


\section{The Proposed Approach}

\begin{figure*}[tb]
\centering
\includegraphics[scale=0.5]{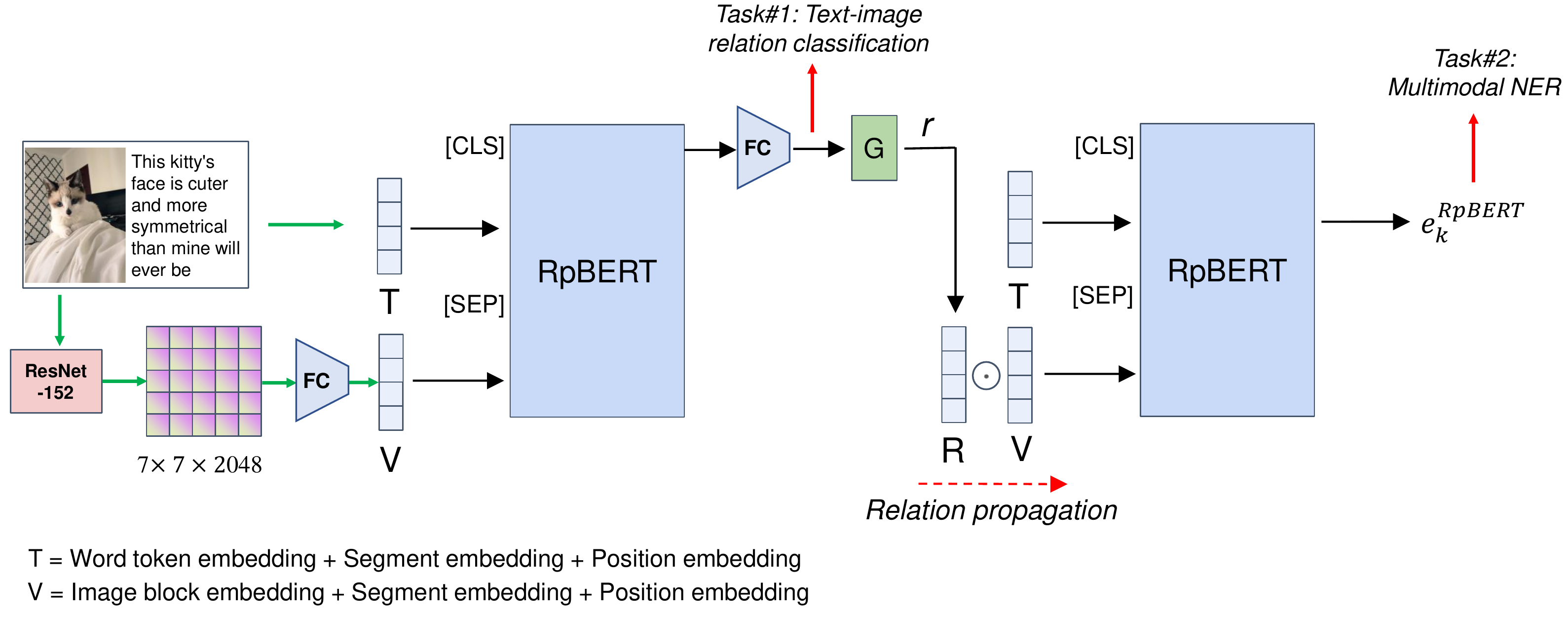}
\caption{The RpBERT architecture overview. Two RpBERTs share the same structure and parameters.}
\label{fig:rpbert}
\end{figure*}

In this section, we introduce a text-image Relation propagation-based BERT model (RpBERT) for multimodal NER, which is shown in Fig.~\ref{fig:rpbert}.
We illustrate the RpBERT architecture and then describe its training procedure in detail.

\subsection{Model Design}
\label{sec:RpBERT}

Our RpBERT extends vanilla BERT to a multitask framework of text-image relation classification and visual-linguistic learning for MNER.
First, similar to most visual-linguistic BERTs, we adapt vanilla BERT to multimodal inputs.
The input sequence of RpBERT is designed as follows:
\begin{equation}
\label{eq:tokens}
\texttt{[CLS]}~\underbrace{w_1~\ldots~w_n}_{\texttt{T}}~\texttt{[SEP]}~\underbrace{v_1~\ldots~v_m}_{\texttt{V}}, 
\end{equation} 
where \texttt{[CLS]} stands for text-image relation classification, \texttt{[SEP]} stands for the separation between text and image features, \texttt{T}=$\{w_1, \ldots, w_n\}$ denotes a sequence of linguistic features, and  \texttt{V}=$\{v_1, \ldots, v_m\}$ denotes a  sequence of visual features.
The word token sequence is generated by the BERT tokenizer, which breaks an unknown word into multiple word-piece tokens.
Unlike the latest visual-linguistic BERT models~\cite{su2019vl,lu2019vilbert,li2020unicoder}, we represent visual features as block regions instead of RoIs.	
The visual features are extracted from the image by ResNet~\cite{he2016deep}.
The output size of the last convolutional layer in ResNet is $7 \times 7\times d_{v}$, where $7 \times 7$ denotes 49 block regions in an image. 
The extracted features of block regions $\{f_{i,j}\}_{i,j=1}^{7}$ are arranged into an image block embedding sequence $\{b_{1}=f_{1,1}W^{v}, \ldots, b_{49}=f_{7,7}W^{v}\}$, where $f_{i,j} \in \mathbb{R}^{1 \times d_v}$ and $W^{v} \in \mathbb{R}^{d_v \times d_{BERT}}$ to match the embedding size of BERT, and $d_{v}=2048$ when working with ResNet-152.
Following the practice in BERT, the input embeddings of tokens are the sum of word token embeddings (or image block embeddings), segment embeddings, and position embeddings.
The  segment embeddings are learned from two types, where A denotes text tokens and  B denotes image blocks.
The  position embeddings of word tokens are learned from the word order in the sentence, but all positions are the same for visual tokens.

The output of the token \texttt{[CLS]} is fed to a fully connected (FC) layer as a binary classifier for Task\#1 of text-image relation classification.
Additionally, we use the probability gate $G$ shown in Fig.~\ref{fig:rpbert} to yield probabilities $[\pi_0,\pi_1]$.
The text-image relevant score $r$ is defined as the probability of being positive,
\begin{equation}
\label{eq:score}
r=\pi_1.
\end{equation} 
We use the relevant score $r$ to construct a visual mask matrix \texttt{R} in Fig.~\ref{fig:rpbert},
\begin{equation}
\label{eq:R}
\texttt{R}= \Big (
	x_{i,j}=r
\Big ) _{49 \times d_{BERT}}.
\end{equation} 
The text-image relation is propagated to  RpBERT via $\texttt{R} \odot \texttt{V}$, where $\odot$ is the element-wise multiplication.
For example, if $\pi_1=0$, then all visual features are discarded.
Finally,  $e^{RpBERT}_k$, the outputs of the tokens \texttt{T} with visual clues, are fed to the NER model  for Task\#2 training.

\subsection{Relation Propagation}

We investigate two kinds of relation propagation, soft and hard, by different probability gates $G$:

\begin{itemize}
\item \textbf{Soft relation propagation}: 
In soft relation propagation, the output of $G$ can be viewed as a continuous distribution. 
The visual features are filtered according to the strength of the text-image relation.
The  gate $G$ is defined as a softmax function:
\begin{equation}
G_{s} = softmax(x).
\end{equation} 

\item \textbf{Hard relation propagation}:  
In hard relation propagation, the output of $G$ can be viewed as a categorical distribution. 
The visual features are either selected or discarded based on 0 or 1.
The gate $G$ is defined as follows:
\begin{equation}
G_{h1} = [softmax(x)>0.5],
\end{equation} 
where $[\cdot]$ is the Iverson bracket indicator function, which takes a value of 1 when its argument is true and 0 otherwise.
As $G_{h1} $ is not differentiable, an empirical way is to use a straight-through estimator~\cite{Bengio13estimating} for propagating  gradients back through the network.
Besides, Jang et al.~\shortcite{jang2017categorical} proposed Gumbel-Softmax to create a continuous approximation to the categorical distribution. 
Inspired by this, we define the gate $G$ as Gumbel-Softmax in Eq.~\eqref{eq:GS} for hard relation propagation.
\begin{equation}
\label{eq:GS}
G_{h2} = softmax((x+g)//\tau),
\end{equation} 
where $g$ is a noise sampled from Gumbel distribution and $\tau$ is a temperature parameter. 
As the temperature approaches 0, samples from the Gumbel-Softmax distribution become one-hot and the Gumbel-Softmax distribution becomes identical to the categorical distribution. In the training stage, the temperature $\tau$ is annealed using the schedule of  1 to 0.1.
\end{itemize}
In the experimental results, we compare the performances of $G_{s}$, $G_{h1}$, and $G_{h2}$  in Table~\ref{tab:CompNER}.

\subsection{Multitask Training for MNER}
\label{sec:tasks}

In this section, we present how to train  RpBERT for MNER.
The training procedure involves multitask leaning of text-image relation classification and MNER, represented by solid red arrows in Fig.~\ref{fig:rpbert}.
The two tasks are described in detail as follows: 

\smallskip
\noindent
\textbf{Task\#1 Text-image relation classification (TRC)}:
We employ the ``Image Task'' splits of the TRC dataset ~\cite{vempala2019categorizing} for text-image relation classification. This classification attempts to identify whether the image’s content contributes additional information beyond the text.
The types of text-image relations and statistics of the TRC dataset are shown in Table~\ref{tab:bloomberg}.
Let $\mathcal D_{1}=\{a^{(i)}\}^{N}_{i=1}=\{<text^{(i)},image^{(i)}>\}^{N}_{i=1}$ be a set of text-image pairs for TRC training. 
The loss $\mathcal L_{1}$ of binary relation classification is calculated by cross entropy:
\begin{equation}
\label{eq:loss1}
\mathcal L_{1} = -\sum^{N}_{i=1} log(p(a^{(i)})),
\end{equation} 
where  $p(x)$ is the probability for correct classification and is calculated by softmax.

\smallskip
\noindent
\textbf{Task\#2  MNER via relation propagation}:
In this stage, we use the mask matrix \texttt{R} to control the additive visual clues.
The input sequence of RpBERT is \texttt{[CLS]}~\texttt{T}~\texttt{[SEP]}~\texttt{R}$\odot$\texttt{V}.
We denote the output of \texttt{T} as $e^{RpBERT}_{k}$.
To perform NER, we use  biLSTM-CRF~\cite{lample2016neural} as a baseline NER model.
The biLSTM-CRF model consists of a bidirectional LSTM and conditional random fields (CRF)~\cite{lafferty2001conditional}.
The  input $e_k$ of biLSTM-CRF is a concatenation of word and character embeddings~\cite{lample2016neural}.
CRF uses the biLSTM hidden vectors of each token to tag the sequence with entity labels.
To evaluate the RpBERT model, we concatenate $e^{RpBERT}_{k}$ as the input of biLSTM, i.e., $[e_k;e^{RpBERT}_{k}]$.
For out-of-vocabulary (OOV) words, we average the outputs of BERT-tokenized subwords not only to generate an approximate vector but also to align the broken words with the input embeddings of biLSTM-CRF.

In biLSTM-CRF, named entity tagging is trained on a standard CRF model.
We feed the hidden vectors $H=\{h_{t}=[\overrightarrow{h}^{LSTM}_{t};\overleftarrow{h}^{LSTM}_{t}]\}^{n}_{t=1}$ of biLSTM to the CRF model.
For a sequence of tags $y= \{y_1,\ldots,y_n\}$, the probability of the label sequence $y$ is defined as follows~\cite{lample2016neural}:
\begin{eqnarray}
\label{eq:CRFprob}
&p(y|x)= \frac{e^{s(x,y)}}{\sum_{y'\in Y} e^{s(x,y')}},
\end{eqnarray}
where $Y$ is all possible tag sequences for the sentence $x$ and $s(x,y)$ are feature functions modeling transitions and emissions.
Details can be referred in~\cite{lample2016neural}.
The objective of Task\#2 is to minimize the negative log-likelihood over the training data $\mathcal D_{2}=\{(x^{(i)},y^{(i)})\}^M_{i=1}$:
 %
\begin{equation}
\label{eq:CRFloss}
\mathcal L_{2} =-\sum^{M}_{i=1} log(p(y^{(i)}|x^{(i)})).
\end{equation} 

\smallskip

Combining Task\#1 and Task\#2, the complete training procedure of RpBERT for MNER is illustrated in Algorithm~\ref{alg:tasks}. 
$\theta_{RpBERT}$, $\theta_{ResNet}$, $\theta_{FCs}$, $\theta_{biLSTM}$, and $\theta_{CRF}$ represent the parameters of RpBERT, ResNet, FCs, biLSTM, and CRF, respectively. 
In each epoch, the procedure first performs Task\#1 to train the text-image relation on the TRC dataset
and then performs Task\#2 to train the model on MNER dataset.
In the test stage, we execute lines 8-10 of Algorithm~\ref{alg:tasks} and decode the valid  sequence of labels using Viterbi algorithm~\cite{lafferty2001conditional}.

\begin{algorithm}[tb]
\caption{Multitask training procedure of RpBERT for MNER.} \label{alg:tasks}
\begin{algorithmic}[1]
\Require The TRC dataset and MNER dataset.
\Ensure  $\theta_{RpBERT}$, $\theta_{ResNet}$, $\theta_{FCs}$, $\theta_{biLSTM}$, and $\theta_{CRF}$.
\For{all epochs}
\For{all batches in the TRC dataset}
\State{Forward text-image pairs through  RpBERT;}
\State{Compute loss $\mathcal L_{1}$ by Eq.~\eqref{eq:loss1};}
\State \multiline{Update $\theta_{FCs}$ and finetune  $\theta_{RpBERT}$ and $\theta_{ResNet}$ using $\nabla {\mathcal L}_{1}$;}
\EndFor
\For{all batches in the MNER dataset}
\State{Forward text-image pairs through  RpBERT;}
\State{Compute the visual mask matrix \texttt{R};}
\State \multiline{Forward text-image pairs with relation propagation through  RpBERT and biLSTM-CRF;}
\State{Compute loss $\mathcal L_2$  by Eq.~\eqref{eq:CRFloss};}
\State \multiline{Update $\theta_{biLSTM}$ and $\theta_{CRF}$ and finetune $\theta_{RpBERT}$ and $\theta_{ResNet}$ using $\nabla {\mathcal L_2}$;}
\EndFor
\EndFor
\end{algorithmic}
\end{algorithm}

\section{Experiments}
\label{sec:length}

\subsection{Datasets}

In the experiments, we use three datasets to evaluate the performance. One is the TRC dataset, and the other two are  MNER  datasets of Fudan University and Snap Research. 
The detailed descriptions are as follows:

\begin{itemize}


\item \textbf{TRC dataset of Bloomberg LP~\cite{vempala2019categorizing}}

In this dataset, the authors annotated tweets into four types of text-image relation, as shown in Table~\ref{tab:bloomberg}.
``Image adds to the tweet meaning'' is centered on the role of the image to the semantics of the tweet while ``Text is presented in image'' focuses on the text’s role.
In the RpBERT model, we treat the  text-image relation for the image’s role as binary classification task between $R_1 \cup R_2$ and $R_3 \cup R_4$.
We follow the same split of 8:2 for train/test sets as in~\cite{vempala2019categorizing}.
We use this dataset to perform learning Task\#1  of  RpBERT.

%

\begin{table}[htb]
\small
\centering
\begin{tabular}{|l|cccc|} 
  \hline
		& $R_1$ & $R_2$   &$R_3$ & 	$R_4$	\\
		 \hline

				Image adds to the tweet meaning				& $\surd$ & $\surd$ & $\times$& $\times$\\
					Text is presented in image			& $\surd$  & $\times$  & $\surd$ &$\times$ \\
					 Percentage (\%) 			& 18.5  &  25.6& 21.9 & 33.8\\
		\hline
\end{tabular}
\caption{Four relation types  in the TRC dataset.}\label{tab:bloomberg}
\end{table}

\item \textbf{MNER dataset of Fudan University~\cite{zhang2018adaptive} }

The authors sampled the tweets with images collected through Twitter’s API.
In this dataset, the NE types are Person, Location, Organization, and Misc.
The authors labeled 8,257 tweet texts using the BIO2 tagging scheme and used a 4,000/1,000/3,257 train/dev/test split.

\item \textbf{MNER  dataset of Snap Research~\cite{lu2018visual}}

The authors collected the data from Twitter and Snapchat, but Snapchat data are not available for public use.
The NE types are Person, Location, Organization, and Misc.
Each data instance contains one sentence and one image.
The authors labeled 6,882 tweet texts using the BIO tagging scheme and used a 4,817/1,032/1,033 train/dev/test split.

\end{itemize}

%

\subsection{Settings}

We use the 300-dimensional fastText Crawl~\cite{mikolov2018advances} word vectors in biLSTM-CRF.
All images are reshaped to a size of $224 \times 224$ to match the input size of ResNet.
We use ResNet-152 to extract visual features and finetune it with a learning rate of 1e-6.
The FC layers in our model are a linear neural network followed by ReLU activation.
The architecture of  RpBERT is the same as  that of BERT-Base, and  we load the pretrained weights from BERT-base-uncased model  to initialize our RpBERT model.
We train the model using Adam~\cite{kingma2014adam} optimizer with default settings. 
Table~\ref{tab:parameters} shows the hyperparameter values in the RpBERT and biLSTM-CRF models.
We use F1 score as evaluation metric for TRC and MNER.


%

\begin{table}[htb]

\begin{center}
\small
\begin{tabular}{|l|c|}

\hline
 Hyperparameter & Value\\
\hline
LSTM hidden state size  &256\\
 ~~~~~+RpBERT& 1024\\
LSTM layer  & 2\\
mini-batch size&8\\
char embedding dimension & 25\\
optimizer& Adam\\
learning rate & 1e-4\\  
learning rate for finetuning RpBERT and ResNet& 1e-6\\  
dropout rate  & 0.5\\
\hline
\end{tabular}
\end{center}
\caption{Hyperparameters of the RpBERT and biLSTM-CRF models.}\label{tab:parameters}
\end{table}

\subsection{Result of TRC}

Table~\ref{tab:CompSemi} shows the performance of  RpBERT in text-image relation classification on the test set of the TRC data.
In terms of the network structure, Lu et al.~\shortcite{lu2018visual} represented the multimodal feature as a concatenation of linguistic features from LSTM and  visual features from InceptionNet~\cite{szegedy2015going}.
The result shows that the BERT-based visual-linguistic model significantly outperforms that of Lu et al.~\shortcite{lu2018visual}, and F1 score of RpBERT on the test set of the TRC data increases by 7.1\% compared to Lu et al.~\shortcite{lu2018visual}.

\begin{table}[htb]

\centering
\small
\begin{tabular}{|l|cc|} 
  \hline
		& Lu et al.~\shortcite{lu2018visual} & RpBERT\\
				\hline
	F1 score&	81.0 & \textbf{(+7.1) 88.1}  \\
						
		\hline
\end{tabular}
\caption{Results of the text-image relation classification in F1 score (\%).}\label{tab:CompSemi}
\end{table}

\subsection{Results of MNER}

Table~\ref{tab:CompNER} illustrates the improved performance by visual clues, such as biLSTM-CRF vs. biLSTM-CRF with image and BERT vs. RpBERT.
The inputs of ``biLSTM-CRF'' and ``biLSTM-CRF + BERT'' are text only, while those of other models are text-image pairs.
``biLSTM-CRF w/ image at $t=0$'' means that the image feature is placed at the beginning of LSTM before the word sequence, similar to the model in \cite{vinyals2015show}.
``biLSTM-CRF + RpBERT'' means that the contextual embeddings $e^{RpBERT}_{k}$ with visual clues are concatenated as the input of  biLSTM-CRF, as clarified in the section of ``Multitask Training for MNER''.
The results show that the best ``+ RpBERT$_{G_s}$'' achieves increases of 4.5\% and 7.3\%  compared to ``biLSTM-CRF'' on the Fudan Univ. and Snap Res. datasets, respectively. 
In terms of the role of visual features, the increase of ``+ RpBERT$_{G_s}$'' achieves approximately 2.5\% compared to ``+ BERT'',
which is larger than those of the biLSTM-CRF based multimodal models such as Zhang et al.~\shortcite{zhang2018adaptive} and Lu et al.~\shortcite{lu2018visual} compared to biLSTM-CRF. 
This indicates that the RpBERT model can better leverage visual features to enhance the context of tweets.

\begin{table}[tb]

\centering
\small
\begin{tabular}{|l|cc|} 
  \hline
		&  Fudan  Univ. &  Snap  Res. \\
 \hline
			biLSTM-CRF & \textbf{(+0.0)} 69.9& \textbf{(+0.0)} 80.1  \\ 
			biLSTM-CRF w/ image at $t=0$ & \textbf{(+0.1)} 70.0& \textbf{(+0.5)} 80.6 \\ 
			Zhang et al.~\shortcite{zhang2018adaptive}  & \textbf{(+0.8)} 70.7 & - \\ 
			Lu et al.~\shortcite{lu2018visual} & - & \textbf{(+0.6)} 80.7 \\ 
			\hline
			biLSTM-CRF + BERT &\textbf{(+0.0)} 71.6& \textbf{(+0.0)} 85.1  \\ 
			biLSTM-CRF + RpBERT$_{G_{h1}}$ &\textbf{(+2.2)} 73.8& (\textbf{+1.3)} 86.4   \\ 
			biLSTM-CRF + RpBERT$_{G_{h2}}$  &\textbf{(+2.6)} 74.2& (\textbf{+1.5)} 86.6   \\ 
			biLSTM-CRF + RpBERT$_{G_{s}}$  &\textbf{(+2.8)} 74.4& (\textbf{+2.3)} 87.4   \\ 
		\hline
\end{tabular}
\caption{Comparison of the improved performance by visual clues in F1 score (\%).}\label{tab:CompNER}
\end{table}

\setcounter{table}{5}

\begin{table*}[tb]

\centering
\small
\begin{tabular}{|l|ccc|ccc|} 

	\hline
		&  \multicolumn{3}{c|}{Fudan Univ.} &  \multicolumn{3}{c|}{Snap Res.} \\\cline{2-7}
				& Image adds &Image doesn't add & Overall & Image adds & Image doesn't add & Overall   \\
				\hline
			biLSTM-CRF + RpBERT$_{G_{s}}$ & 74.6  & 74.1  & 74.4 & 87.7  & 86.9 & 87.4\\			
		  ~~~~~~~~~~- w/o Rp &\textbf{(-0.5)} 74.1& \textbf{(-3.1)} 71.0& \textbf{(-1.8)} 72.6 & \textbf{(-0.7)} 87.0& \textbf{(-2.3)}  84.6 & \textbf{(-1.2)} 86.2 \\
		\hline
\end{tabular}
\caption{Performance comparison in F1 score (\%) when the relation propagation (Rp) is ablated.}\label{tab:RGNAblated}
\end{table*}

\setcounter{table}{4}

\begin{table}[tb]
\centering
\small
\begin{tabular}{|l|cc|} 

	 \hline
		&Fudan Univ.&Snap Res.\\
	\hline
		Arshad et al.~\shortcite{arshad2019aiding}& 72.9  & -  \\
			Yu et al.~\shortcite{yu2020improving}& 73.4	 & 85.3  \\	
			biLSTM-CRF + VL-BERT & 72.4&   86.0 \\ 
		  biLSTM-CRF + ViLBERT & 72.0&  85.7  \\ 
			biLSTM-CRF + UNITER & 72.7& 86.1\\
			biLSTM-CRF + RpBERT$_{G_s}$ & 74.4  &  87.4  \\
		  biLSTM-CRF + RpBERT-Large$_{G_s}$ & \textbf{74.9}&  \textbf{87.8}  \\	
								\hline
\end{tabular}
\caption{Performance comparison with other models in F1 score (\%).}\label{tab:embedding}
\end{table}

\setcounter{table}{6}
In Table~\ref{tab:embedding}, we compare performance with the state-of-the-art method~\cite{yu2020improving} and visual-linguistic pretrained models which codes are available, such as VL-BERT~\cite{su2019vl}, ViLBERT~\cite{lu2019vilbert}, and UNITER~\cite{chen2020uniter}.
Similar to $e^{RpBERT}_{k}$ in RpBERT, we take out the contextual embeddings of word sequence in visual-linguistic models
and concatenate them with the token embeddings $e_{k}$ as the input embedding of biLSTM-CRF.
For example, ``biLSTM-CRF + VL-BERT'' means that the output of word sequence in VL-BERT is concatenated as the input of  biLSTM-CRF, i.e., $\left[e_k;e^{VL\mbox{-}BERT}_{k}\right]$.
The results show that RpBERT$_{G_s}$ outperforms all pretrained models.
Additionally,  we test RpBERT using the structure of BERT-Large, which has 24 layers and 16 attention heads.
``biLSTM-CRF + RpBERT-Large$_{G_s}$'' achieves state-of-the-art results on the MNER datasets and outperforms the current best results~\cite{yu2020improving} by 1.5\% on the Fudan Univ. dataset and 2.5\% on the Snap Res. dataset. 

\begin{table*}[!h]
	\centering
	\small
	\begin{tabular}{|m{0.14\textwidth}|m{0.14\textwidth}|m{0.14\textwidth}|m{0.14\textwidth}|m{0.14\textwidth}|m{0.14\textwidth}|}
		\hline
		&\centering   1	& \centering   2 & \centering  3 &  \centering  4 &  \centering\arraybackslash  5  \\
		\hline
		 \centering Image& \centering \begin{minipage}[m]{0.14\textwidth}
			\includegraphics[width=1.0\textwidth]{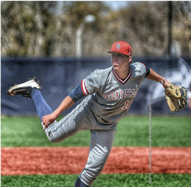}
		\end{minipage}
		& \centering \begin{minipage}[m]{0.14\textwidth}
				\includegraphics[width=1.0\textwidth]{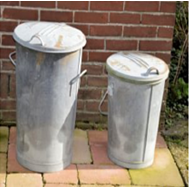}
		\end{minipage}
		&\centering  \begin{minipage}[m]{0.14\textwidth}
				\includegraphics[width=1.0\textwidth]{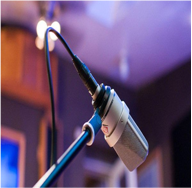}
		\end{minipage}
				&\centering  \begin{minipage}[m]{0.14\textwidth}
				\includegraphics[width=1.0\textwidth]{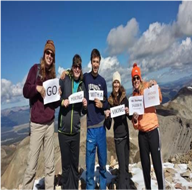}
		\end{minipage}
		&\centering\arraybackslash  \begin{minipage}[m]{0.14\textwidth}
				\includegraphics[width=1.0\textwidth]{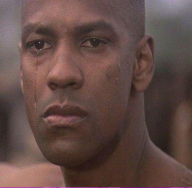}
		\end{minipage}\\
		
		%

		\hline
	\centering	$\{a^v_j\}^{49}_{j=1}$ of RpBERT w/o Rp & \centering \begin{minipage}[m]{0.14\textwidth}
			\includegraphics[width=1.0\textwidth]{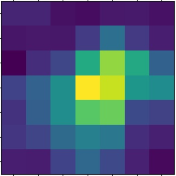}
		\end{minipage}	
		& \centering \begin{minipage}[m]{0.14\textwidth}
			\includegraphics[width=1.0\textwidth]{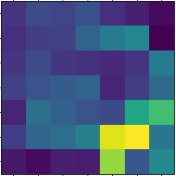}
		\end{minipage}	
		&\centering \begin{minipage}[m]{0.14\textwidth}
			\includegraphics[width=1.0\textwidth]{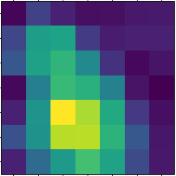}
		\end{minipage}
		&\centering \begin{minipage}[m]{0.14\textwidth}
			\includegraphics[width=1.0\textwidth]{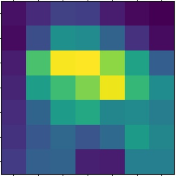}
		\end{minipage}
		&\centering \arraybackslash \begin{minipage}[m]{0.14\textwidth}
			\includegraphics[width=1.0\textwidth]{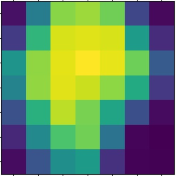}
		\end{minipage} \\
		\hline
		
	\centering	$\{a^v_j\}^{49}_{j=1}$ of RpBERT$_{G_s}$  & \centering \begin{minipage}[m]{0.14\textwidth}
			\includegraphics[width=1.0\textwidth]{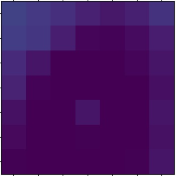}
		\end{minipage}	
		& \centering \begin{minipage}[m]{0.14\textwidth}
			\includegraphics[width=1.0\textwidth]{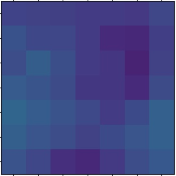}
		\end{minipage}	
		&\centering \begin{minipage}[m]{0.14\textwidth}
			\includegraphics[width=1.0\textwidth]{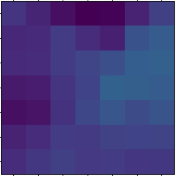}
		\end{minipage}
				&\centering \begin{minipage}[m]{0.14\textwidth}
			\includegraphics[width=1.0\textwidth]{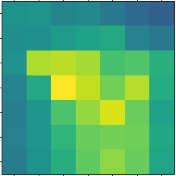}
		\end{minipage}
		&\centering \arraybackslash \begin{minipage}[m]{0.14\textwidth}
			\includegraphics[width=1.0\textwidth]{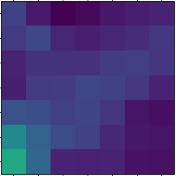}
		\end{minipage} \\
		\hline
	 	 \centering   $r$& 
		\centering 0.14 &  \centering 0.13 & \centering 0.24 & \centering 0.74 & \centering\arraybackslash  0.20 \\
		\hline
	\centering	+ RpBERT w/o Rp & Looking forward to editing some {\color{blue}\textbf{[ORG SBU]}} baseball shots from Saturday.
		& Nice image of \textbf{[PER Kevin Love]} and \textbf{[PER Kyle Korver]} during 1 st half \# NBAFinals \# Cavsin9 \# {\color{red}\textbf{[LOC Cleveland]}}
		& {\color{red}\textbf{[ORG Reddit]}} needs to stop pretending racism is valuable debate.
		& {\color{red}\textbf{[MISC PSD] Lesher}} teachers take school spirit to top of 14ner  \color{blue}\textbf{[LOC Mount Sherman]}.
		& Ask {\color{red}\textbf{[PER Siri]}} what 0 divided by 0 is and watch her put you in your place.\\
		\hline
\centering	 + RpBERT$_{G_s}$& Looking forward to editing some {\color{blue}\textbf{[ORG SBU]}} baseball shots from Saturday.
		& Nice image of \textbf{[PER Kevin Love]} and \textbf{[PER Kyle Korver]} during 1 st half \# NBAFinals \# Cavsin9 \# \color{blue}\textbf{[ORG Cleveland]}
		& {\color{blue}\textbf{[MISC Reddit]}} needs to stop pretending racism is valuable debate.
		& {\color{blue}\textbf{[ORG PSD Lesher]}} teachers take school spirit to top of 14ner  \color{blue}\textbf{[LOC Mount Sherman]}.
			&Ask {\color{blue}\textbf{[MISC Siri]}} what 0 divided by 0 is and watch her put you in your place.\\
		\hline
\centering	Previous work
		& Looking forward to editing some \textbf{{\color{red}SBU}} baseball shots from Saturday. \textit{\cite{lu2018visual}}
		& Nice image of \textbf{[PER Kevin Love]} and \textbf{[PER Kyle Korver]} during 1 st half \# NBAFinals \# Cavsin9 \# {\color{red}\textbf{[LOC Cleveland]}}. \textit{\cite{lu2018visual}}
		&  {\color{red}\textbf{[ORG Reddit]}} needs to stop pretending racism is valuable debate. \textit{\cite{arshad2019aiding}}
		&  {\color{blue}\textbf{[ORG PSD Lesher]}} teachers take school spirit to top of 14ner  {\color{red}\textbf{[PER Mount Sherman]}}. \textit{\cite{arshad2019aiding}}
					& Ask {\color{red}\textbf{[PER Siri]}} what 0 divided by 0 is and watch her put you in your place. \textit{\cite{yu2020improving}}\\
		\hline
		
		\multicolumn{5}{c}{		 \hspace{7em}  Low ~~~~~~\begin{minipage}[p]{0.29\textwidth}
				\vspace{0.6em} 	\includegraphics[scale=0.8]{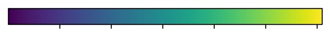}
			\end{minipage} High}\\
		
	\end{tabular}
		\caption{Five failed examples in the previous works tested by  ``+ RpBERT$_{G_s}$'' and ``+ RpBERT w/o Rp''. Blue and black labels are correct and red ones are wrong.}\label{tab:case}
\end{table*}

%

\subsection{Ablation Study}


In this section, we report the results when ablating the relation propagation in  RpBERT,
or equivalently  performing only Task\#2  in the training of  RpBERT.
Table~\ref{tab:RGNAblated} shows that the overall performance without relation propagation (``w/o Rp'') decreases by -1.8\% and -1.2\% on the Fudan Univ. and Snap Res. datasets, respectively.
In addition, we divide the test data into two sets, ``Image adds'' and ``Image doesn't add'', by the text-image relation classification, and compare the impact of the ablation on the data of different relation types.
The performances on all relation types are improved with relation propagation.
More importantly, regarding the ``Image doesn't add'' type, ``w/o Rp'' lowers the F1 scores by a large margin, -3.1\% on the Fudan Univ. dataset and -2.3\% on the Snap Res. dataset.
This justifies that the text-unrelated visual features exert large negative effects on learning visual-linguistic representations.

\begin{figure}[tb]
\centering
\begin{subfigure}{0.23\textwidth}
	\centering
	\includegraphics[width=1.05\linewidth]{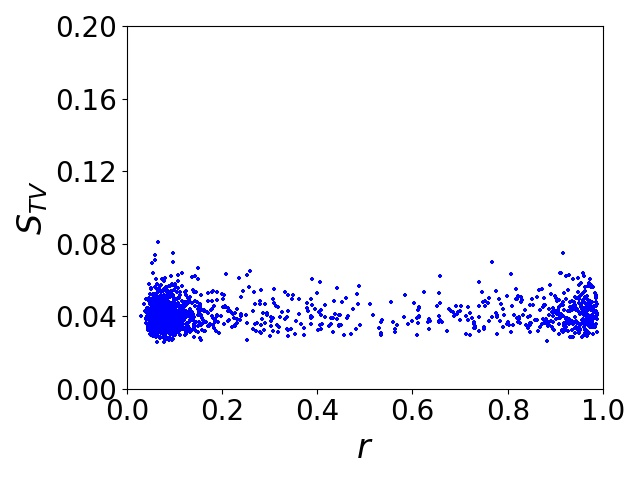}
	\caption{RpBERT w/o Rp}
\end{subfigure}
\begin{subfigure}{0.23\textwidth}
	\centering
	\includegraphics[width=1.05\linewidth]{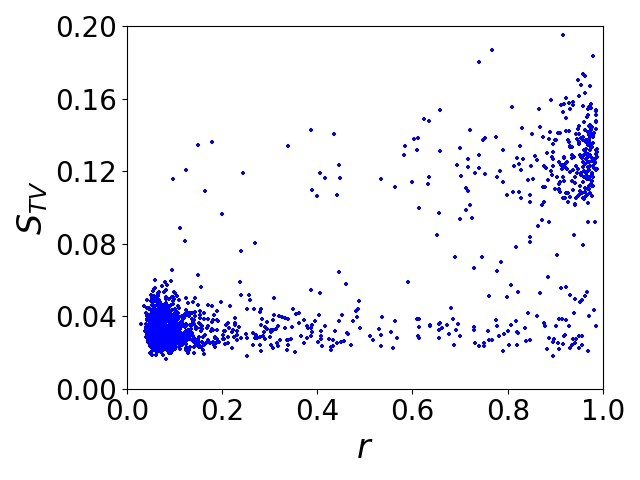}
	\caption{RpBERT$_{G_s}$}
\end{subfigure}
\caption{The numerical distribution between $r$ and $S_{TV}$.}
\label{fig:scorev}
\end{figure}

In Fig.~\ref{fig:scorev}, we illustrate the comparison of RpBERT and RpBERT w/o Rp in terms of the numerical distribution between the relevant score $r$ and $S_{TV}$, where $S_{TV}$ is the average sum of visual attentions and is defined as follows:
\begin{equation}
S_{TV} = \frac{1}{LH}\sum^{L}_{l=1}\sum^{H}_{h=1}\sum^{n}_{i=1}\sum^{m}_{j=1}Att^{(l,h)}(w_i,v_j),
\end{equation}
where $Att^{(l,h)}(w_i,v_j)$ is the attention between the $i$th word and $j$th image block on the $h$th head and $l$th layer in RpBERT.
The samples are from the test set of the Snap Res. dataset.
In Fig.~\ref{fig:scorev}(a), we find that the distribution of $S_{TV}$ of RpBERT w/o Rp is close to a horizontal line and is unrelated to the relevant score $r$.
In Fig.~\ref{fig:scorev}(b), most $S_{TV}$ values of RpBERT decrease on irrelevant text-image pairs ($r<0.5$) and increase on relevant text-image pairs ($r>0.5$) compared to those of RpBERT w/o Rp.
Quantitatively, the mean of  $S_{TV}$  decreases by 20\% from 0.041 to 0.034 on irrelevant text-image pairs 
while it increases from 0.042 to  0.102 on relevant text-image pairs.
In general, after using relation propagation, the trend is towards leveraging more visual cues in stronger text-image relations.


\subsection{Case Study via Attention Visualization}
We illustrate five failure examples mentioned in \cite{lu2018visual,arshad2019aiding,yu2020improving} in Table~\ref{tab:case}.
The common reason for these failed examples is inappropriate visual attention features.
The table shows the relevant score $r$ and overall image attentions of RpBERT and RpBERT w/o Rp.
The visual attention of an image block $j$ across all words, heads and layers is defined as follows:
\begin{equation}
a^v_j=\frac{1}{LH}\sum^{L}_{l=1}\sum^{H}_{h=1}\sum^{n}_{i=1}Att^{(l,h)}(w_i,v_j).
\end{equation}
We visualize the overall image attentions $\{a^v_j\}^{49}_{j=1}$ by heat maps.
The NER results of ``+ RpBERT w/o Rp'', ``+ RpBERT$_{G_s}$'', and the previous works are also presented for comparison.

Examples 1 and 2 are from the Snap Res. dataset, and Examples 3, 4, and 5 are from the Fudan Univ. dataset.
The NER results of all examples obtained by  RpBERT are correct.
In Example 1,  RpBERT performs correct  and the visual attentions have no negative effects on the NER results.   
In Example 2, the visual attentions focus on the ground and result in  tagging ``Cleveland'' with the wrong label ``LOC''.
In Example 3, ``Reddit'' is misidentified as ``ORG'' by  the visual attentions.
In Example 5, ``Siri'' is wrongly identified as ``PER'' because of  the visual attentions to the human face.
In Examples 2, 3, and 5, the text-image pairs are recognized as irrelevant since the values of $r$ are small.
With text-image relation propagation, much less visual features are weighted to the linguistic features in RpBERT and the NER results are  correct.
In Example 4, the text and image are related, i.e., $r=0.74$.
The persons are significantly concerned in \cite{arshad2019aiding}, resulting in the wrong label ``PER'' for ``Mount Sherman''.
RpBERT w/o Rp extends some visual attention to the mountain scene,
while RpBERT increases much more visual attention to the scenery, such as sky and mountain, and thus strengthens the understanding of the whole picture and yields the correct labels of ``PSD Lesher'' and ``Mount Sherman''.

%

\section{Conclusion}
This paper concerns the visual attention problem raised by the text-unrelated images in tweets for multimodal learning.
We propose a relation propagation-based multimodal model based on text-image relation inference.
The model is trained by the multiple tasks of  text-image relation classification and downstream NER.
%
In experiments, the ablation study quantitatively evaluates the role of text-image relation propagation.
The heat map visualization and  numerical distribution regarding the visual attention justify that RpBERT can better leverage visual information adaptively according to the relation between text and image.
The failed cases mentioned in other papers are effectively resolved by the RpBERT model.
Our model achieves the best F1 scores in both TRC and MNER, i.e., 88.1\% on the TRC dataset, 74.9\% on the Fudan Univ. dataset, and 87.8\% on the Snap Res. dataset.
%

\section{Acknowledgements}
This work was supported by the National Innovation and Entrepreneurship Training Program for College Students under Grant 202013021005 
and in part by the National Natural Science Foundation of China (NSFC) under Grant 62072402.
%

\bibliography{bibTex}

\begin{thebibliography}{29}
\providecommand{\natexlab}[1]{#1}
\providecommand{\url}[1]{\texttt{#1}}
\providecommand{\urlprefix}{URL }
\expandafter\ifx\csname urlstyle\endcsname\relax
  \providecommand{\doi}[1]{doi:\discretionary{}{}{}#1}\else
  \providecommand{\doi}{doi:\discretionary{}{}{}\begingroup
  \urlstyle{rm}\Url}\fi

\bibitem[{Arshad et~al.(2019)Arshad, Gallo, Nawaz, and
  Calefati}]{arshad2019aiding}
Arshad, O.; Gallo, I.; Nawaz, S.; and Calefati, A. 2019.
\newblock Aiding Intra-Text Representations with Visual Context for Multimodal
  Named Entity Recognition.
\newblock In \emph{2019 International Conference on Document Analysis and
  Recognition (ICDAR)}, 337--342.

\bibitem[{Bengio, L{\'{e}}onard, and Courville(2013)}]{Bengio13estimating}
Bengio, Y.; L{\'{e}}onard, N.; and Courville, A.~C. 2013.
\newblock Estimating or Propagating Gradients Through Stochastic Neurons for
  Conditional Computation.
\newblock \emph{arXiv preprint arXiv:1308.3432} .

\bibitem[{Chen et~al.(2015)Chen, Fang, Lin, Vedantam, Gupta, Doll{\'a}r, and
  Zitnick}]{chen2015microsoft}
Chen, X.; Fang, H.; Lin, T.-Y.; Vedantam, R.; Gupta, S.; Doll{\'a}r, P.; and
  Zitnick, C.~L. 2015.
\newblock Microsoft coco captions: Data collection and evaluation server.
\newblock \emph{arXiv preprint arXiv:1504.00325} .

\bibitem[{Chen et~al.(2020)Chen, Li, Yu, El~Kholy, Ahmed, Gan, Cheng, and
  Liu}]{chen2020uniter}
Chen, Y.-C.; Li, L.; Yu, L.; El~Kholy, A.; Ahmed, F.; Gan, Z.; Cheng, Y.; and
  Liu, J. 2020.
\newblock UNITER: UNiversal Image-TExt Representation Learning.
\newblock In \emph{Computer Vision -- ECCV 2020}, 104--120.

\bibitem[{Girshick(2015)}]{girshick2015fast}
Girshick, R. 2015.
\newblock Fast r-cnn.
\newblock In \emph{Proceedings of the IEEE international conference on computer
  vision}, 1440--1448.

\bibitem[{He et~al.(2016)He, Zhang, Ren, and Sun}]{he2016deep}
He, K.; Zhang, X.; Ren, S.; and Sun, J. 2016.
\newblock Deep residual learning for image recognition.
\newblock In \emph{Proceedings of the IEEE conference on computer vision and
  pattern recognition}, 770--778.

\bibitem[{Hosseini(2019)}]{hosseini2019implicit}
Hosseini, H. 2019.
\newblock Implicit entity recognition, classification and linking in tweets.
\newblock In \emph{Proceedings of the 42nd International ACM SIGIR Conference
  on Research and Development in Information Retrieval}, 1448--1448.

\bibitem[{Hu et~al.(2017)Hu, Zheng, Yang, and Huang}]{hu2017twitter100k}
Hu, Y.; Zheng, L.; Yang, Y.; and Huang, Y. 2017.
\newblock Twitter100k: A real-world dataset for weakly supervised cross-media
  retrieval.
\newblock \emph{IEEE Transactions on Multimedia} 20(4): 927--938.

\bibitem[{Jang, Gu, and Poole(2017)}]{jang2017categorical}
Jang, E.; Gu, S.; and Poole, B. 2017.
\newblock Categorical Reparameterization with Gumbel-Softmax.
\newblock In \emph{International conference on learning representations}.

\bibitem[{Kingma and Ba(2014)}]{kingma2014adam}
Kingma, D.~P.; and Ba, J. 2014.
\newblock Adam: A method for stochastic optimization.
\newblock \emph{arXiv preprint arXiv:1412.6980} .

\bibitem[{Lafferty, McCallum, and Pereira(2001)}]{lafferty2001conditional}
Lafferty, J.~D.; McCallum, A.; and Pereira, F.~C. 2001.
\newblock Conditional Random Fields: Probabilistic Models for Segmenting and
  Labeling Sequence Data.
\newblock In \emph{Proceedings of the Eighteenth International Conference on
  Machine Learning}, 282--289.

\bibitem[{Lample et~al.(2016)Lample, Ballesteros, Subramanian, Kawakami, and
  Dyer}]{lample2016neural}
Lample, G.; Ballesteros, M.; Subramanian, S.; Kawakami, K.; and Dyer, C. 2016.
\newblock Neural Architectures for Named Entity Recognition.
\newblock In \emph{Proceedings of the 2016 Conference of NAACL-HLT}, 260--270.
  Association for Computational Linguistics.

\bibitem[{Li et~al.(2020)Li, Duan, Fang, Gong, Jiang, and
  Zhou}]{li2020unicoder}
Li, G.; Duan, N.; Fang, Y.; Gong, M.; Jiang, D.; and Zhou, M. 2020.
\newblock Unicoder-VL: A Universal Encoder for Vision and Language by
  Cross-Modal Pre-Training.
\newblock In \emph{AAAI}, 11336--11344.

\bibitem[{Li et~al.(2019)Li, Yatskar, Yin, Hsieh, and Chang}]{li2019visualbert}
Li, L.~H.; Yatskar, M.; Yin, D.; Hsieh, C.-J.; and Chang, K.-W. 2019.
\newblock Visualbert: A simple and performant baseline for vision and language.
\newblock \emph{arXiv preprint arXiv:1908.03557} .

\bibitem[{Lu et~al.(2018)Lu, Neves, Carvalho, Zhang, and Ji}]{lu2018visual}
Lu, D.; Neves, L.; Carvalho, V.; Zhang, N.; and Ji, H. 2018.
\newblock Visual attention model for name tagging in multimodal social media.
\newblock In \emph{Proceedings of the 56th Annual Meeting of the Association
  for Computational Linguistics (Volume 1: Long Papers)}, 1990--1999.

\bibitem[{Lu et~al.(2019)Lu, Batra, Parikh, and Lee}]{lu2019vilbert}
Lu, J.; Batra, D.; Parikh, D.; and Lee, S. 2019.
\newblock Vilbert: Pretraining task-agnostic visiolinguistic representations
  for vision-and-language tasks.
\newblock In \emph{Advances in Neural Information Processing Systems}, 13--23.

\bibitem[{Mikolov et~al.(2018)Mikolov, Grave, Bojanowski, Puhrsch, and
  Joulin}]{mikolov2018advances}
Mikolov, T.; Grave, E.; Bojanowski, P.; Puhrsch, C.; and Joulin, A. 2018.
\newblock Advances in Pre-Training Distributed Word Representations.
\newblock In \emph{Proceedings of the International Conference on Language
  Resources and Evaluation (LREC 2018)}.

\bibitem[{Moon, Neves, and Carvalho(2018)}]{moon2018multimodal}
Moon, S.; Neves, L.; and Carvalho, V. 2018.
\newblock Multimodal Named Entity Recognition for Short Social Media Posts.
\newblock In \emph{Proceedings of the 2018 Conference of the North American
  Chapter of the Association for Computational Linguistics: Human Language
  Technologies, Volume 1 (Long Papers)}, 852--860.

\bibitem[{Sharma et~al.(2018)Sharma, Ding, Goodman, and
  Soricut}]{sharma-etal-2018-conceptual}
Sharma, P.; Ding, N.; Goodman, S.; and Soricut, R. 2018.
\newblock Conceptual Captions: A Cleaned, Hypernymed, Image Alt-text Dataset
  For Automatic Image Captioning.
\newblock In \emph{Proceedings of the 56th Annual Meeting of the Association
  for Computational Linguistics (Volume 1: Long Papers)}, 2556--2565.
  Melbourne, Australia: Association for Computational Linguistics.

\bibitem[{Simonyan and Zisserman(2014)}]{simonyan2014very}
Simonyan, K.; and Zisserman, A. 2014.
\newblock Very deep convolutional networks for large-scale image recognition.
\newblock \emph{arXiv preprint arXiv:1409.1556} .

\bibitem[{Su et~al.(2019)Su, Zhu, Cao, Li, Lu, Wei, and Dai}]{su2019vl}
Su, W.; Zhu, X.; Cao, Y.; Li, B.; Lu, L.; Wei, F.; and Dai, J. 2019.
\newblock VL-BERT: Pre-training of Generic Visual-Linguistic Representations.
\newblock In \emph{International Conference on Learning Representations}.

\bibitem[{{Szegedy} et~al.(2015){Szegedy}, {Wei Liu}, {Yangqing Jia},
  {Sermanet}, {Reed}, {Anguelov}, {Erhan}, {Vanhoucke}, and
  {Rabinovich}}]{szegedy2015going}
{Szegedy}, C.; {Wei Liu}; {Yangqing Jia}; {Sermanet}, P.; {Reed}, S.;
  {Anguelov}, D.; {Erhan}, D.; {Vanhoucke}, V.; and {Rabinovich}, A. 2015.
\newblock Going deeper with convolutions.
\newblock In \emph{2015 IEEE Conference on Computer Vision and Pattern
  Recognition (CVPR)}, 1--9.

\bibitem[{Tan and Bansal(2019)}]{tan2019lxmert}
Tan, H.; and Bansal, M. 2019.
\newblock LXMERT: Learning Cross-Modality Encoder Representations from
  Transformers.
\newblock In \emph{Proceedings of the 2019 Conference on Empirical Methods in
  Natural Language Processing and the 9th International Joint Conference on
  Natural Language Processing (EMNLP-IJCNLP)}, 5103--5114.

\bibitem[{Vempala and Preo{\c{t}}iuc-Pietro(2019)}]{vempala2019categorizing}
Vempala, A.; and Preo{\c{t}}iuc-Pietro, D. 2019.
\newblock Categorizing and Inferring the Relationship between the Text and
  Image of Twitter Posts.
\newblock In \emph{Proceedings of the 57th Annual Meeting of the Association
  for Computational Linguistics}, 2830--2840.

\bibitem[{Vinyals et~al.(2015)Vinyals, Toshev, Bengio, and
  Erhan}]{vinyals2015show}
Vinyals, O.; Toshev, A.; Bengio, S.; and Erhan, D. 2015.
\newblock Show and tell: A neural image caption generator.
\newblock In \emph{Proceedings of the IEEE conference on computer vision and
  pattern recognition}, 3156--3164.

\bibitem[{Wang, Deyu, and He(2019)}]{wang2019open}
Wang, R.; Deyu, Z.; and He, Y. 2019.
\newblock Open Event Extraction from Online Text using a Generative Adversarial
  Network.
\newblock In \emph{Proceedings of the 2019 Conference on Empirical Methods in
  Natural Language Processing and the 9th International Joint Conference on
  Natural Language Processing (EMNLP-IJCNLP)}, 282--291.

\bibitem[{Yu and Jiang(2019)}]{yu2019adapting}
Yu, J.; and Jiang, J. 2019.
\newblock Adapting BERT for target-oriented multimodal sentiment
  classification.
\newblock In \emph{Proceedings of the 28th International Joint Conference on
  Artificial Intelligence}, 5408--5414. AAAI Press.

\bibitem[{Yu et~al.(2020)Yu, Jiang, Yang, and Xia}]{yu2020improving}
Yu, J.; Jiang, J.; Yang, L.; and Xia, R. 2020.
\newblock Improving Multimodal Named Entity Recognition via Entity Span
  Detection with Unified Multimodal Transformer.
\newblock In \emph{Proceedings of the 58th Annual Meeting of the Association
  for Computational Linguistics}, 3342--3352.

\bibitem[{Zhang et~al.(2018)Zhang, Fu, Liu, and Huang}]{zhang2018adaptive}
Zhang, Q.; Fu, J.; Liu, X.; and Huang, X. 2018.
\newblock Adaptive co-attention network for named entity recognition in tweets.
\newblock In \emph{Thirty-Second AAAI Conference on Artificial Intelligence}.

\end{thebibliography}

\end{document}